\documentclass{ieeeaccess}

\usepackage[dvipsnames]{xcolor}
\usepackage{booktabs}
\usepackage{tikz}
  \NewSpotColorSpace{PANTONE}
\AddSpotColor{PANTONE} {PANTONE3015C} {PANTONE\SpotSpace 3015\SpotSpace C} {1 0.3 0 0.2}
\SetPageColorSpace{PANTONE}
\usepackage{array} 	
\usepackage{booktabs}
\usepackage{dcolumn}
\usepackage{multirow}
\usepackage{prettyref}
\usepackage{amssymb}
\usepackage{amsmath} 
\usepackage{amsfonts}
\usepackage{bbm}
\usepackage{mathtools} 
\usepackage{blindtext}
\usepackage{xspace}
\usepackage{prettyref}
\usepackage{float}
\usepackage{pdflscape}
\usepackage{rotating}
\usepackage{pgfplots}
\usepackage{amssymb}
\usepackage{pgfplots}
\usepackage{pgfplotstable}
\usepgfplotslibrary{units,groupplots}
\pgfplotsset{compat=newest, 
	ylabsh/.style={every axis y label/.style={at={(0,0.5)}, xshift=#1, rotate=90}}} 

\usetikzlibrary{calc,trees,positioning,arrows,chains,shapes.geometric,%
	decorations.pathreplacing,decorations.pathmorphing,shapes,%
	matrix,shapes.symbols}

\usepackage{booktabs}
\usepackage{eurosym}
\usepackage{array}
\usepackage{tabularx}
\newcolumntype{L}{>{\raggedright\arraybackslash}X}

\newrefformat{fig}{fig. \ref{#1}}
\newrefformat{Fig}{Fig. \ref{#1}}
\newrefformat{tab}{tab. \ref{#1}}
\newrefformat{Tab}{Tab. \ref{#1}}
\newrefformat{sec}{section \ref{#1}}
 \usepackage{url}
 \usepackage{array}
 \newcolumntype{P}[1]{>{\centering\arraybackslash}p{#1}}
 \newcolumntype{M}[1]{>{\centering\arraybackslash}m{#1}}
 
\newcommand{\specialcell}[2][c]{
	\begin{tabular}[#1]{@{}c@{}}#2\end{tabular}}

\def\BibTeX{{\rm B\kern-.05em{\sc i\kern-.025em b}\kern-.08em
    T\kern-.1667em\lower.7ex\hbox{E}\kern-.125emX}}
\begin{document}
	
\history{Date of publication xxxx 00, 0000, date of current version xxxx 00, 0000.}
\doi{10.1109/ACCESS.2017.DOI}

\title{Stress Testing Method for Scenario-Based Testing of Automated Driving Systems}

\author{\uppercase{Demin Nalic}\authorrefmark{1}, \IEEEmembership{Member, IEEE},
	\uppercase{Hexuan Li}\authorrefmark{1}, \uppercase{Arno Eichberger}\authorrefmark{1}, \uppercase{Christoph Wellershaus}\authorrefmark{1},\uppercase{Aleksa Pandurevic}\authorrefmark{1} and \uppercase{Branko Rogic}\authorrefmark{2}.}
\address[1]{Institute of Automotive Engineering - Graz University of Technology, Graz, Austria}
\address[2]{MAGNA Steyr Fahrzeugtechnik AG Co \& KG, Graz, Austria}
\tfootnote{This work has been submitted to the IEEE for possible publication. 
	Copyright may be transferred without notice, after which this version may no 
	longer be accessible.}

\markboth
{Demin Nalic \headeretal: Preparation of Papers for IEEE TRANSACTIONS and JOURNALS}
{Demin Nalic \headeretal: Preparation of Papers for IEEE TRANSACTIONS and JOURNALS}

\corresp{Corresponding author: Demin Nalic (e-mail: demin.nalic@tugraz.at).}

\begin{abstract}
	Classical approaches for testing of automated driving systems (ADS) of SAE levels 1 and 2 were based on defined scenarios with specific maneuvers, depending on the function under test. For ADS of SAE level 3+, the scenario space is infinite and calling for virtual testing and verification. The biggest challenge for virtual testing methods lies in the realistic representation of the virtual environment where the ADS is tested. Such an environment shall provide the possibility to model and develop vehicles, objects, control algorithms, traffic participants and environment elements in order to generate valid and representative test data. An important and crucial aspect of such environments is the testing of vehicles in a complex traffic environment with a stochastic and realistic traffic representation. For this research we used a microscopic traffic flow simulation software (TFSS) PTV Vissim and the vehicle simulation software IPG CarMaker to test ADS. Although the TFSS provides realistic and stochastic behavior of traffic participants, the occurrence of safety-critical scenarios (SCS) is not guaranteed. To generate and increase such scenarios, a novel stress testing method (STM) is introduced. With this method, traffic participants are manipulated in the vicinity of the vehicle under test in order to provoke SCS derived from statistical accident data on motorways in Austria. Using the co-simulation between IPG CarMaker, PTV Vissim and external driver models in Vissim are used to imitate human driving errors, resulting in an increase of SCS.
\end{abstract}


\titlepgskip=-15pt

\maketitle

\section{INTRODUCTION}\label{sec:01_introduction}

Testing of ADS using TFSS is an inevitable part of virtual testing procedures. The demand on test kilometers emphasized in the works of \cite{padock}-\cite{winner} and the simulation-based testing in complex virtual environments with objects, traffic, pedestrians, etc., is needed for valid and representative simulation results. The literature in virtual testing is mainly focused on scenario-based testing presented in \cite{survey1}-\cite{critical_1}, where SCS are based on a-priori knowledge coming from accident research, field operational test and expert knowledge. However, if a new system with high complexity is introduced, new failures could emerge. The presented STM inherently includes new SCS based on stochastic virtual testing. New failure patterns related to the specific ADS function are detected and could even serve as a source of knowledge for scenario-based testing. In \cite{ftgFramework}, a framework for testing of ADS including a calibrated traffic flow model (TFM) for testing and generation of SCS on motorways has been presented. This simulation framework combines a vehicle simulation software IPG CarMaker with the TFSS in Vissim. This framework provides a calibrated traffic flow model (TFM) and a possible testing environment to verify the results of the STM. The TFM is behavior-based and the movement is stochastic and unpredictable with regards to the paths and decisions of traffic participants. As for this paper, we focus on motorways, SCS (e.g. collisions, near-collisions or accidents) have not occurred frequently in the used co-simulation. The reason is the lack of errors in vehicle guidance executed by TFSS, whereas those occasionally occur in manually driven vehicles. Therefore, the presented STM is used to increase the number of SCS for virtual scenario-based testing where TFM are considered in the simulation e.g. in \cite{TrafficFlowTesting1}-\cite{TrafficFlowTesting3}. To avoid this lack of errors, statistical accident data from Austrian motorways provided from \cite{statisticAustria} were examined for accident types that occur most frequently. These accident types were classified to longitudinal and lateral accidents and used for STM. Based on this examination, the traffic participants are manipulated to provoke SCS in the vicinity of the vehicle under test. This can be done by using the external driving model interface of Vissim described in \cite{Vissim} and \cite{TrafficFlow}. The interface provides the possibility to implement driver models with defined driving behavior using external driver models implemented by a dynamic link library (DLL) provided in Vissim. \\
The stress testing approach in stochastic virtual ADS testing is a development tool addressing many issues related to automated vehicle guidance, such as velocity control in ACC and AEB, LKA and automated lane change but also more complex issues such as time delays \cite{Wang} and actuator failures as well as network attacks \cite{Deng}. The verification of the STM method will be examined by comparing simulation results with and without using the STM.\\

\section{SIMULATION ENVIRONMENT}\label{sec:02_traffic_flow}

\subsection{Co-Simulation between CarMaker and Vissim}

The simulation environment is based on the co-simulation framework presented in \cite{ftgFramework} and \cite{ifacFramework}. This framework combines Matlab/SIMULINK and a co-simulation between IPG CarMaker and Vissim. In this context, IPG CarMaker is used for the implementation of the automated driving functions and sensor models for simulating machine perception as well as the visualization of simulation test runs. Additionally, it provides a complex multi-body vehicle model for a detailed representation of the ego-vehicle dynamics, as it is necessary in the non-linear region such as emergency braking or skidding. The traffic is generated by Vissim which uses a microscopic, behavioral and time step-oriented traffic flow simulation model introduced in \cite{Vissim} and \cite{trapp}. The term "microscopically" refers to each driver-vehicle unit which can be modelled individually and therefore each vehicle has an individual driving behavior by assigning a specific set of behavior-related parameters in the driver models of each of the vehicle unit driving through the road network. For this paper, a calibrated TFM from \cite{ftgFramework} was used. The calibration is based on traffic measurements on macro- and microscopic level on the official test road for automated driving in Austria called ALP.Lab. The description and introduction to this test road is presented in \cite{alplab}.

\subsection{External Driver DLL in Vissim}\label{sec:externalDLL}
The manipulation of the surrounding vehicles for the STM is carried out by replacing internal driving behavior models with user-defined models. The DLL software framework implemented in C++ is presented in detail in \cite{deminDLL}. In order to achieve this, defined scenarios described in the sections \ref{sec:brakingBeh} and \ref{sec:latBehavior} are implemented in a DLL. During the simulation, Vissim calls the DLL code for each affected vehicle in each simulation time-step to determine the behavior of the vehicle.
%

\section{Stress Testing Method}\label{sec:STM}

\subsection{Accident Data Base}\label{sec:statistic}

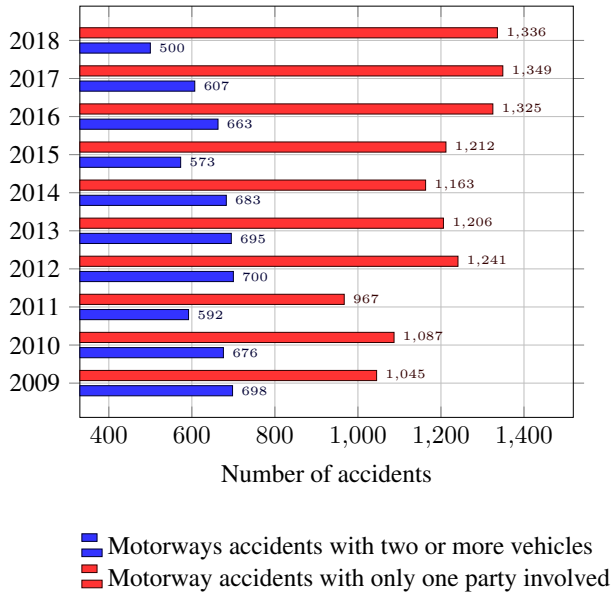
\begin{figure}
\begin{tikzpicture}[scale=0.95]
  \begin{axis}[
    xbar,
    y axis line style = { opacity = 1 },
    x axis line style = { opacity = 1 },
    enlarge y limits  = 0.1 ,
    enlarge x limits  = 0.2,    
    ytick  = {data},
    bar width		  = 4pt,
    grid,
    symbolic y coords = {2009, 2010, 2011, 2012,2013, 2014, 2015, 2016, 2017, 2018},
    nodes near coords,
    every node near coord/.append style={font=\tiny},
    nodes near coords align={horizontal},
    legend cell align={left},
    legend style={anchor=south,
    legend pos= south east,
    yshift = -2.7cm, 
    xshift = 0.9cm,
    xlabel = Number of accidents,
    draw=none
    }
  ]
  \addplot[blue!20!black,fill=blue!80!white] coordinates { 
  						 (698,2009) (676,2010)
  						 (592,2011) (700,2012)
  						 (695,2013) (683,2014)
  						 (573,2015) (663,2016)
  						 (607,2017) (500,2018)                     
                         };
  \addplot[red!20!black,fill=red!80!white] coordinates { 
  						 (1045,2009) (1087,2010) 
  						 (967,2011)  (1241,2012)
  						 (1206,2013) (1163,2014) 
  						 (1212,2015) (1325,2016)
  						 (1349,2017) (1336,2018)                     
                         };
  \legend{Motorways accidents with two or more vehicles, Motorway accidents with only one party involved}
  \end{axis}
\end{tikzpicture}
\caption{Comparison of Accidents with only one party involved and accidents with two or more participants}
\label{fig:accidentsAustria}
\end{figure}

In order to provoke representative stress situations, a statistical database of accident data from the Austrian motorways was used to define maneuvers which are provoked in the surrounding area of the vehicle under test. Accident data has been provided by Statistics Austria and contains all accident data between 2009 and 2018. The classification of motorway accidents as defined by Statistics Austria can be found in \cite{statisticAustriaAccidentClassification}. In this paper, we focus on those classifications that are most likely to occur. Two very dominant accident classifications with their accident types are shown in Tab. \ref{tab:clusterAccident}.
\begin{table}
	\begin{center}
		\begin{tabular}{cl}
			\hline
			\hline
			\multicolumn{2}{|c|}{\textbf{Accidents with only one party involved}}                                   \\ \hline
			\multicolumn{1}{|c|}{Type} & \multicolumn{1}{c|}{Description}                                          \\ \hline
			\multicolumn{1}{|c|}{1}    & \multicolumn{1}{l|}{Leaving the road on the right}                        \\ \hline
			\multicolumn{1}{|c|}{2}    & \multicolumn{1}{l|}{Leaving the road on the left}                         \\ \hline
			\multicolumn{1}{|c|}{3}    & \multicolumn{1}{l|}{Leaving the road by a lane intersection or road exit} \\ \hline
			\multicolumn{1}{|c|}{4}    & \multicolumn{1}{l|}{Reverse driving or turn around}                       \\ \hline
			\multicolumn{1}{|c|}{5}    & \multicolumn{1}{l|}{Downfall of or in the vehicle}                        \\ \hline
			\multicolumn{1}{|c|}{6}    & \multicolumn{1}{l|}{Collision with an object}                             \\ \hline
			\multicolumn{1}{|c|}{7}    & \multicolumn{1}{l|}{Other accidents with only one party involved}         \\ \hline
			\multicolumn{2}{l}{}                                                                                   \\ \hline
			\multicolumn{2}{|c|}{\textbf{Accidents with two or more parties involved}}                   \\ \hline
			\multicolumn{1}{|c|}{Type} & \multicolumn{1}{c|}{Description}                                          \\ \hline
			\multicolumn{1}{|c|}{1}    & \multicolumn{1}{l|}{Collision during overtaking}                          \\ \hline
			\multicolumn{1}{|c|}{2}    & \multicolumn{1}{l|}{Lane changing with and without collision}             \\ \hline
			\multicolumn{1}{|c|}{3}    & \multicolumn{1}{l|}{Collision with a moving vehicle}                      \\ \hline
			\multicolumn{1}{|c|}{4}    & \multicolumn{1}{l|}{Collision with a stationary vehicle due to traffic}   \\ \hline
			\multicolumn{1}{|c|}{5}    & \multicolumn{1}{l|}{Collision at an intersection}                         \\ \hline
			\multicolumn{1}{|c|}{6}    & \multicolumn{1}{l|}{Collision due to reverse driving}                     \\ \hline
			\multicolumn{1}{|c|}{7}    & \multicolumn{1}{l|}{Collision due to get into the lane}                   \\ \hline
			\multicolumn{1}{|c|}{8}    & \multicolumn{1}{l|}{Other accidents with one-way traffic}                 \\ \hline \hline
		\end{tabular}
		\caption{Two accident classes with the defined types of the accident.}
		\label{tab:clusterAccident}	
\end{center}
\end{table}
%
The total number of accidents caused by the types presented in Tab. \ref{tab:clusterAccident} is shown in Fig. \ref{fig:accidentsAustria}. These two accident classes, accidents with one party involved and accidents with two or more parties involved, are the most frequent on Austrian motorways and they comprise 94\% of all motorway accidents from 2009 until 2018.
%
From Fig. \ref{fig:accidentsAustria} it is obvious that traffic accidents where one party is involved are significantly higher and therefore, they were handled in more detail.
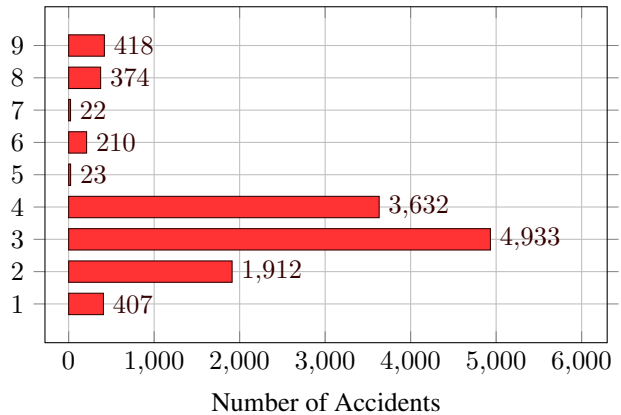
\begin{figure}
\begin{tikzpicture}
  \begin{axis}[  
  	height = 6cm,
  	width = 9cm,
    xbar,
    y axis line style = { opacity = 1 },
    x axis line style = { opacity = 1 },
    enlarge y limits  = 0.15 ,
    enlarge x limits  = 0.05,
    xmax = 6000,
    grid,
    ytick  			  = data,
    ytick style		  = { opacity = 1 },
    bar width		  =8pt,
    nodes near coords,
    legend cell align={left},
    legend style={anchor=south,
    legend pos= south east,
    yshift = -1.0cm, 
    xshift = 0,
	xlabel = Number of Accidents,  
    draw=none
    }
  ]
 \addplot[red!20!black,fill=red!80!white] 
 			coordinates {(	(407,1)
							(1912,2)
							(4933,3)    
							(3632,4)    
							(23,5)
 							(210,6)    
 							(22,7)
 							(374,8)
							(418,9)  
};                      
 \end{axis}
\end{tikzpicture}

\caption{Accidents with one party involved on Austrian motorways with the clusters: (1) Collision during overtaking (2) Lane change with and without collision (3) Collision with a moving vehicle (4) Collision with a stationary vehicle due to traffic (5) Collision at an intersection (6) Collision at an intersection (7) Collision due to reverse driving (8) Collision due to get into the lane (9) Other accidents with one-way traffic}
\label{fig:oneWayAccident}
\end{figure}
%
In Fig. \ref{fig:oneWayAccident}, the clusters of traffic accidents with one party involved on Austrian motorways are depicted. Based on these clusters, the four most frequent accidents types are chosen and are relevant for the STM. By abstracting these accident types in longitudinal and lateral types we define four relevant scenarios for STM, which are shown in Tab. \ref{tab:accidents}.
\begin{table}
	\begin{tabular}{|l|>{\centering\arraybackslash}p{1.5cm}|>{\centering\arraybackslash}p{1.5cm}|}
		\hline
		\hline
		   & \textbf{Lateral}           & \textbf{Longitudinal}          \\ \hline
		Collision with a moving vehicle     & x                    & - \\ \hline
		Collision with a stationary vehicle & x                    & - \\ \hline
		Lane Change with collision          & - & x                    \\ \hline
		Lane Change without collision       & - & x                    \\ 
		\hline
		\hline
	\end{tabular}
\caption{Relevant accident types used for the STM.}
\label{tab:accidents}
\end{table}
%

\subsection{Test Method Description}
\begin{figure} 
	\centering
	\scalebox{0.75}{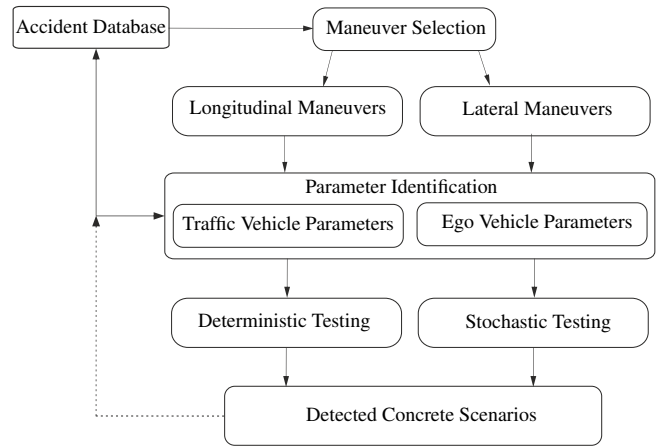}
	\caption{Stress testing method procedure.}
	\label{fig:stmProcedure} 
\end{figure}

The general procedure of the STM is depicted in Fig. \ref{fig:stmProcedure}. Figure \ref{fig:stmProcedure} shows that outputs of the STM are concrete scenarios which could complement the accident database and can be used for parameter identification as in \cite{parIdent1} and \cite{parIdent2}. Concrete scenarios are well described in \cite{concreteScenario}. The parameter identification and methods for concrete scenario generation are not part of this paper. The traffic participants of the surrounding area of the vehicle under test are manipulated to provoke scenarios using maneuvers based on scenarios from Tab. \ref{tab:accidents}. This maneuvers are called stress test events (STE).
%

\subsection{Stress Testing for Longitudinal Scenarios}

\begin{figure}[H]
	\centering
	\scalebox{0.5}{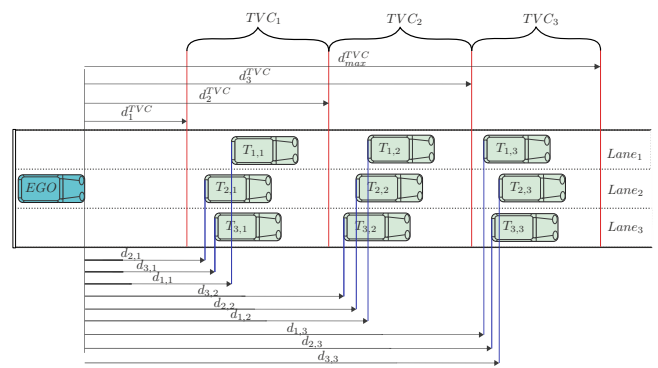}
	\caption{Relevant traffic vehicle the 3 lane configuration of the STM.}
	\label{fig:relevantCars}
\end{figure}

In order to provoke STE in longitudinal direction, traffic vehicles $T_{i,j}$ as depicted in Fig. \ref{fig:relevantCars} will be manipulated in a defined manner. 
%
The index $i \in\{1,2,3\}$ represents the considered traffic vehicle column (TVC) and index  $j \in\{1, 2, 3\}$ the lane number of the traffic vehicle. These two index definitions are used in further equations and matrices. According to Fig. \ref{fig:relevantCars} the TVCs represent the locations of actual traffic vehicles in front of the EGO vehicle within a defined distance interval and 3 columns. 
The manipulation process for longitudinal STEs is carried out in such a way that the vehicle decelerates as described in section \ref{sec:brakingBeh}. The distances to each $d_{k}^{TVC}$ with $k \in\{1, 2, 3, max\}$ depend on the safety interval times (SIT) $t_{k}^{s}$ and current EGO vehicle speed $v_{ego}$ and are calculated as in (\ref{eq:eq1}). The dependence on the EGO vehicle speed makes them variable during the simulation and adjustable with the SIT $t_{k}^{s}$.
\begin{equation}
d_{k}^{TVC} = v_{ego} t_{k}^{s}, \:\:k\in\{1, 2, 3, max\}\label{eq:eq1}
\end{equation}

%

In Fig. \ref{fig:relevantCars} a 3-lane traffic configuration is presented, the same procedure is defined for 2-lane roads. Mandatory for the STM is the definition of distance matrices $\textbf{D}^{l^{2}}$ for 2 and $\textbf{D}^{l^{3}}$ for 3 lanes. The upper indices $l^{2}$ and $l^{3}$ of the matrices represent highways sections with 2 lanes and 3 lanes, respectively. The matrices are defined in (\ref{eq:eq2}).
\begin{equation}
	\begin{split}
		\textbf{D}^{l^{2}} &= \begin{bmatrix} d_{1,1} &  d_{1,2}&  d_{1,3}\\  d_{2,1} &  d_{2,2} & d_{2,3}\end{bmatrix} \in \rm I\!R^{2x3}\\
		\textbf{D}^{l^{3}} &= \begin{bmatrix} d_{1,1} &  d_{1,2}&  d_{1,3}\\  d_{2,1} &  d_{2,2} & d_{2,3}\\  d_{3,1} &  d_{3,2} & d_{3,3} \end{bmatrix} \in \rm I\!R^{3x3}\label{eq:eq2}
	\end{split}
\end{equation}
%
They represent the distance from the EGO vehicle to each target vehicle in the TVC column and they are calculated as in (\ref{eq:eq3}).
\begin{equation}
	d_{i,j}=s_{i,j}^{T}-s_{ego} \label{eq:eq3}
\end{equation}
%
In (\ref{eq:eq3}) $s_{ego}$ represents the EGO vehicle position and $s_{i,j}^{T}$ the position of the traffic vehicle. The distance matrices are calculated continuously during the simulation and are used for further steps of the STM.\\
To implement the STM in Vissim using the described external driver DLL, the steps from Tab. \ref{tab:stmSteps} are defined and implemented. In the context of this work an event represents a maneuver which is provoked to bring the EGO vehicle in a stress condition. How these events are defined is described in the following steps.\\
\begin{table}
	\begin{center}
	\begin{tabular}{ |c | l|}
		\hline
		\hline 
		\textbf{Step 1} & Traffic event matrix (TEM)\\ 
		\textbf{Step 2} & Event trigger conditions\\
		\textbf{Step 3} & Event counter\\ 
		\textbf{Step 4} & Scenario time frame\\    
		\textbf{Step 5} & Storage of vehicle states\\ 
		\hline
		\hline
	\end{tabular}
	\caption{Definitions of required implementation steps for STM.}
	\label{tab:stmSteps}
	\vspace{-0.5cm}
\end{center}
\end{table}
%
\textbf{Step 1}: Traffic Event Matrix (TEM)\\
Depending on the number of road lanes we differentiate between two TEM which are defined in (\ref{eq:tem}).
\begin{equation}\label{eq:tem}
	\begin{split}
		\textbf{E}_{T}^{l^{2}} &= \begin{bmatrix} e_{1,1}^{l^{2}} &  e_{1,2}^{l^{2}} &  e_{1,3}^{l^{2}}\\  e_{2,1}^{l^{2}} &  e_{2,3}^{l^{2}} & e_{2,3}^{l^{2}} \end{bmatrix} \in \rm I\!R^{2x3}\\
		\textbf{E}_{T}^{l^{3}} &= \begin{bmatrix} e_{1,1}^{l^{3}} &  e_{1,2}^{l^{3}} &  e_{1,3}^{l^{3}}\\  e_{2,1}^{l^{3}} &  e_{2,3}^{l^{3}} & e_{2,3}^{l^{3}}\\  e_{3,1}^{l^{3}} &  e_{3,3}^{l^{3}} & e_{3,3}^{l^{3}} \end{bmatrix} \in \rm I\!R^{3x3}
	\end{split}
\end{equation}
%
The matrix coefficients $e_{i,j}^{l^{2}}$ and $e_{i,j}^{l^{3}}$ are Boolean values and they define whether a car is in a certain TVC or not, see Fig. \ref{fig:relevantCars}. The calculation of the matrix entries $e_{i,j}^{l^{2}}$ and $e_{i,j}^{l^{3}}$ in (\ref{eq:eq5}) is done by checking if the entries of distances matrices $\textbf{D}^{l_{2}}$ and $\textbf{D}^{l_{3}}$ in (\ref{eq:eq2}) are in the range of distances $d_{k}^{TVC}$.
\begin{equation}
e_{i,j}^{l^{2}} = e_{i,j}^{l^{3}} =
\begin{cases}
\text{1,} &\quad d_{1}^{TVC} < d_{i,j} < d_{2}^{TVC} \\
\text{1,} &\quad d_{2}^{TVC} < d_{i,j} < d_{3}^{TVC} \\
\text{1,} &\quad d_{3}^{TVC} < d_{i,j} < d_{max}^{TVC} \\
\text{0,} &\quad otherwise\\
\end{cases} \label{eq:eq5}
\end{equation}
%
\begin{figure}[H]
	\centering
	\scalebox{0.8}{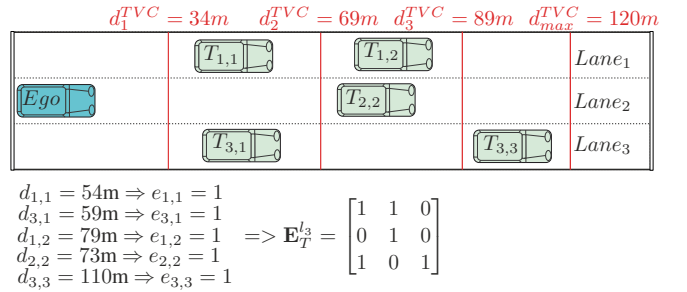}
	\caption{Example of calculating a TEM using a random traffic event.}	
	\label{fig:temDescription}
\end{figure} 
%
An example how of a random traffic event in Fig. \ref{fig:temDescription} shows how the TEM is calculated. In the simulation the TEM is calculated continuously until a trigger condition is executed.


\begin{table}
	\begin{center}
		\renewcommand{\arraystretch}{1.5}
		\begin{tabular}{|p{0.1\textwidth}|p{0.3\textwidth}|}
			\hline
			\hline
			\textbf{Requirement 1} & Triggering of target vehicles $T_{i,j}$ can only be carried out in TVC column.                                                                                                                                                                                \\ \hline
			\textbf{Requirement 2} & The manipulation of traffic participants is triggered if the target vehicles $T_{i,j}$ a laying on the same lane as the ego vehicle, no matter on which TVC it occur.                                                                                         \\ \hline
			\textbf{Requirement 3} & More then one target vehicles can only be triggered if the second or third target vehicle are in the same TVC column.                                                                                                                                         \\ \hline
			\textbf{Requirement 4} & The triggered longitudinal STE start first with the triggering of vehicles in the first $TVC_{1}$ column, then the second $TVC_{2}$ and then third $TVC_{3}$ column. After reaching a certain number of events in one column, these events are not triggered anymore. \\ \hline \hline
		\end{tabular}%
	\end{center}
	\caption{Requirement for the combination matrices $\textbf{C}_{q}^{l_{2}}$ and $\textbf{C}_{q}^{l_{3}}$.}
	\label{tab:requirements}
	\vspace{-0.5cm}
\end{table}
%
\textbf{Step 2}: Event Trigger Conditions\\
As we consider longitudinal stress maneuvers in this chapter, the events which are triggered are braking maneuvers (BM). These BM are divided in two braking categories which could be triggered for the STM. The detailed definition of the BM is described in section \ref{sec:brakingBeh}.\\ Based on the TEM defined in (\ref{eq:tem}) the combination matrices are defined in (\ref{eq:compMatrix}).

\begin{equation}	
	\label{eq:compMatrix}
	\begin{split}
		\textbf{C}_{q}^{l_{2}} &= \begin{bmatrix} c_{1,1}^{l_{2}} &  c_{1,2}^{l_{2}} &  c_{1,3}^{l_{2}}\\  c_{2,1}^{l_{2}} &  c_{2,2}^{l_{2}} & c_{2,3}^{l_{2}} \end{bmatrix} \in \rm I\!R^{2x3}\\
		\textbf{C}_{q}^{l_{3}} &= \begin{bmatrix} c_{1,1}^{l_{3}} &  c_{1,2}^{l_{3}} &  c_{1,3}^{l_{3}}\\  c_{2,1}^{l_{3}} &  c_{2,2}^{l_{3}} & c_{2,3}^{l_{3}}\\  c_{3,1}^{l_{3}} &  c_{3,3}^{l_{3}} & c_{3,3}^{l_{3}} \end{bmatrix} \in \rm I\!R^{3x3}
	\end{split}
\end{equation}
%
The matrices $\textbf{C}_{q}^{l_{2}}$ and $\textbf{C}_{q}^{l_{3}}$ represent relevant traffic conditions which should be tested for two and three traffic lanes, respectively. The index $q$ represents the number of relevant braking combinations. For the triggering of braking maneuvers, the combination matrices in Tab. \ref{tab:comp1} and Tab. \ref{tab:comp2} are relevant for the STM. Entry 1 means the car was on this position and an event has occurred, so the car should start braking.
\begin{table}
	\begin{center}
		\small\addtolength{\tabcolsep}{-2pt}
		\begin{tabular}{|c | c c c c c c| }
			\hline
			\hline\\[-0.8em]
			   	 		   		      & $c_{1,1}^{l_{2}}$ &  $c_{1,2}^{l_{2}}$ &  $c_{1,3}^{l_{2}}$ & $c_{2,1}^{l_{2}}$ &  $c_{2,3}^{l_{2}}$ & $c_{2,3}^{l_{2}}$\\
			 \hline\\[-1em]
			 $\textbf{C}_{1}^{l_{2}}$ & 1 & X & X & X & X & X\\
			 \hline\\[-1em]
			 $\textbf{C}_{2}^{l_{2}}$ & 1 & X & X & 1 & X & X\\
			 \hline\\[-1em]
			 $\textbf{C}_{3}^{l_{2}}$ & X & 1 & X & X & X & X\\
			 \hline\\[-1em]
			 $\textbf{C}_{4}^{l_{2}}$ & X & 1 & X & X & 1 & X\\ 
			 \hline\\[-1em]
			 $\textbf{C}_{5}^{l_{2}}$ & X & X & 1 & X & X & X\\
			 \hline\\[-1em]
			 $\textbf{C}_{6}^{l_{2}}$ & X & X & 1 & X & X & 1\\  
			 \hline\\[-1em]
			 $\textbf{C}_{7}^{l_{2}}$ & X & X & X & 1 & X & X\\
			 \hline\\[-1em]
			 $\textbf{C}_{8}^{l_{2}}$ & X & X & X & X & 1 & X\\
			 \hline\\[-1em]
			 $\textbf{C}_{9}^{l_{2}}$ & X & X & X & X & X & 1\\
			 \hline
			 \hline
		\end{tabular}
		\caption{Comparison matrix entries for 2 lane highway. \hspace{\textwidth}}
		\label{tab:comp1}
	\end{center}
\vspace{-1cm}
\end{table}
%
\begin{table}
	\begin{center}
		\small\addtolength{\tabcolsep}{-2pt}
		\begin{tabular}{|c | c c c c c c c c c|}
			\hline
			\hline\\[-0.8em]
		  	& $c_{1,1}^{l_{3}}$ &  $c_{1,2}^{l_{3}}$ &  $c_{1,3}^{l_{3}}$ & $c_{2,1}^{l_{3}}$ &  $c_{2,2}^{l_{3}}$ & $c_{2,3}^{l_{3}}$ & $c_{3,1}^{l_{3}}$ &  $c_{3,2}^{l_{3}}$ & $c_{3,3}^{l_{3}}$\\
			\hline\\[-1em]
			 $\textbf{C}_{1}^{l_{3}}$ & 1 & X & X & 1 & X & X & 1 & X & X\\
			 \hline\\[-1em]
			 $\textbf{C}_{2}^{l_{3}}$& X & 1 & X & X & 1 & X & X & 1 & X\\
			 \hline\\[-1em]
			 $\textbf{C}_{3}^{l_{3}}$ & X & X & 1 & X & X & 1 & X & X & 1\\ 
			 \hline\\[-1em]
			 $\textbf{C}_{4}^{l_{3}}$ & 1 & X & X & X & X & X & X & X & X\\
			 \hline\\[-1em]
			 $\textbf{C}_{5}^{l_{3}}$ & X & X & X & 1 & X & X & X & X & X\\
			 \hline\\[-1em]
			 $\textbf{C}_{6}^{l_{3}}$ & X & X & 1 & X & X & X & 1 & X & X\\ 
			 \hline\\[-1em]
			 $\textbf{C}_{7}^{l_{3}}$ & X & 1 & X & X & X & X & X & X & X\\
			 \hline\\[-1em]
			 $\textbf{C}_{8}^{l_{3}}$ & X & X & X & X & 1 & X & X & X & X\\
			 \hline\\[-1em]
			 $\textbf{C}_{9}^{l_{3}}$ & X & X & 1 & X & X & X & X & 1 & X\\ 
			 \hline\\[-1em]
			 $\textbf{C}_{10}^{l_{3}}$ & X & X & 1 & X & X & X & X & X & X\\
			 \hline\\[-1em]
			 $\textbf{C}_{11}^{l_{3}}$ & X & X & X & X & X & 1 & X & X & X\\
			 \hline\\[-1em]
			 $\textbf{C}_{12}^{l_{3}}$ & X & X & X & X & X & X & X & X & 1\\  
			\hline
			\hline
		\end{tabular}
		\caption{Comparison matrix entries for 3 lane highway. \hspace{\textwidth}}
		\label{tab:comp2}
	\end{center}
\vspace{-1cm}
\end{table}
%
Entry X could be 0 or 1 but the vehicles on this position will not brake. According to these tables, the combination matrices are examined by satisfying the requirements form Tab. \ref{tab:requirements}. The braking maneuvers will be triggered only if in at least one TVC column a traffic vehicle occur and only in a situation when the traffic vehicle is located in the same lane as the ego vehicle. By comparing the event matrices $\textbf{E}_{T}^{l_{2}}$ or $\textbf{E}_{T}^{l_{3}}$ with the combination matrices $\textbf{C}_{q}^{l_{2}}$ and $\textbf{C}_{q}^{l_{3}}$ an event is triggered if the matrix entries match. The event shown in Fig. \ref{fig:temDescription} would yield to the matrix entries shown in (\ref{eq:eq7}).
\begin{equation}
	\begin{split}
		\textbf{E}_{T}^{l_{3}} &= \begin{bmatrix} 1 &  \textbf{1} &  0\\  0 &  \textbf{1} & 0\\  1 &  0 & 1 \end{bmatrix} \\
		\textbf{C}_{q}^{l_{3}} &= \begin{bmatrix} X & \textbf{1}  &  X\\  X &  \textbf{1} & X\\  X &  X & X \end{bmatrix} \\
		e_{1,2} &= e_{2,2} = c_{1,2} = c_{2,2} 
	\end{split}\label{eq:eq7}
\end{equation}
%
If such a event occur, the trigger time $t_{trigger}$ is saved and an event counters $n_{ct}^{l_{2}}$ and $n_{ct}^{l_{3}}$ are incremented.


\textbf{Step 3}: Event Counter\\
Each event which occurs is counted by incrementing the trigger counter $n_{ct}^{l_{2}}$ or $n_{ct}^{l_{3}}$. To make sure that all events happen at least a certain number of times a maximum number of occurrences $n_{ct}^{max}$ during a simulation run is defined and is used as parametrization parameter of the STM. That means, if a certain combination from the matrices $\textbf{C}_{q}^{l_{2}}$ or $\textbf{C}_{q}^{l_{3}}$ occur $n_{ct}^{max}$-th times they will not be triggered any more during a simulation run.


\textbf{Step 4}: Scenario Time Frame\\
Each detected scenario by the STM is saved within a certain time frame. This time frame is defined by the trigger time $t_{trigger}$, the upper frame limit $t_{upper}$ and lower frame limit $t_{lower}$. The upper- and lower-time frame limits can be adjusted and together with the trigger time $t_{trigger}$ they represent the scenario time frame $t_{scene}$ shown in (\ref{eq:eq8}).
\begin{equation}
	t_{trigger} - t_{lower} < t_{scene} < t_{trigger} + t_{upper} \label{eq:eq8}
\end{equation}
%


\textbf{Step 5}: Storage of Vehicle States\\

In case that a scenario is detected all data listed in Tab. \ref{tab:trafficData} will be saved for further analysis. The stored data will contain all vehicles in the scenario time frame $t_{scene}$. This includes only the relevant vehicles depicted in Fig. \ref{fig:relevantCars}.
\begin{table}[H]
	\begin{center}
		\begin{tabular}{ |l| l|}
			\hline
			\hline
			$s_{ego}$ & Position of the ego vehicle\\ 
			$v_{ego}$ & Velocity of the ego vehicle\\
			$a_{ego}$ & Acceleration of the ego vehicle\\ 
			$s_{i,j}^{T}$ & Position of all traffic vehicles\\    
			$v_{i,j}^{T}$ & Velocity of all traffic vehicles\\
			$a_{i,j}^{T}$ & Acceleration of all traffic vehicles\\
			$d_{i,j}$ & All entries of the $\textbf{D}^{l_{2}}$ and $\textbf{D}^{l_{3}}$ matrices\\ 
			distance & Driven distance of the ego and traffic vehicles\\    
			time & Simulation time\\
			\hline
			\hline
		\end{tabular}
		\caption{Vehicle states and parameters which are saved if a STE occur.}
		\label{tab:trafficData}
	\end{center}
\end{table}
%


\subsection{Acceleration and Deceleration Behaviour}\label{sec:brakingBeh}

In Tab. \ref{tab:accidents}, longitudinal scenarios with a stationary vehicle in front are relevant scenarios for the STM. These types of accidents can be caused by full braking maneuvers or sudden speed reductions on the motorway. The type of accidents where a vehicle is approaching an existing crash will not be considered since this type of accidents is exhausted in various research and industry works in the field of active safety systems. For more details about active safety systems, see \cite{activeSafety1}-\cite{activeSafety5}. Considering that, braking behavior of vehicles during the simulation has a significant impact on the validity of the results of the STM. As described in section \ref{sec:externalDLL}, we use the external driver DLL in Vissim to redesign the deceleration characteristics, which will be more effective to improve the realistic simulation of the vehicle. Majority of studies in the past have proposed deceleration models, Akçelik and Biggs suggested non-uniform deceleration rate to describe a polynomial behavior between acceleration and speed \cite{Akcelik1}. However, since previous models were limited to the study of cars and trucks only, dual regime models depict the deceleration behavior of various vehicle types that were proposed in  \cite{Maurya} and \cite{Bokare}. Some studies demonstrate that deceleration profile can be presented using a polynomial fit which shows that the maximum deceleration typically occurs during the braking stabilization phase on ESP-equipped cars \cite{Kudarauskas}, therefore, we can simplify the deceleration profile into a polynomial model. As described above, previous studies have established that the basic deceleration profiles can be fitted in a polynomial model, thus the deceleration for different traffic scenarios needs to be discussed as well. 
\begin{table} [H]
	\begin{center}
		{
			\renewcommand{\arraystretch}{1.5}
			\begin{tabular}{|p{0.1\textwidth}|p{0.3\textwidth}|}
				\hline
				\hline
				Equation & Polynomial model is proposed by Akçelik and Biggs \cite{Akcelik2}\\
				\hline
				
				Deceleration & The deceleration is divided into two intervals to represent the different driving behavior, the comfort deceleration interval is defined as [0, -3.5] m/s$^{2}$. The driver's deceleration interval becomes [-3.5, -8.5] m/s$^{2}$ in some emergency situations are described in \cite{P}\cite{Mehmood} .\\
				\hline
				
				Braking time & Based on research experiments presented in \cite{Wang2} the braking time are in the range between 10.1s and 17.2s\\
				\hline
				
				Braking distance & The average braking distance for different speed intervals is presented in \cite{Wang2}.\\
				\hline
				\hline
				
			\end{tabular}
			}
		\caption{Driver braking deceleration model. \hspace{\textwidth}}
	\label{tab:DriverBrakingDec}
	\end{center}
\vspace{-1cm}
\end{table}
%

\begin{table}[H]
	\begin{center}
		\renewcommand{\arraystretch}{1.5}
		\begin{tabular}{|p{0.1\textwidth}|p{0.3\textwidth}|}
			\hline
			\hline
			Equation & Parabolic model\\
			\hline
			
			Deceleration & Based on ISO 22179 \cite{ISO22179} the average automatic deceleration shall not exceed 3.5 m/s$^2$ when the vehicle velocity above 20 m/s and 5 m/s$^2$ (average value over 2s) when the vehicle velocity is below 5 m/s.\\
			\hline
			
			Deceleration gradient & The relationships between deceleration, jerk value and speed are specified by the ISO 22179 in \cite{ISO22179}. Those specifications are considered for the calculation of the deceleration gradients for the STM.\\
			\hline
			\hline
			
		\end{tabular}
		\caption{ACC braking deceleration model. \hspace{\textwidth}}
		\label{tab:ACCrBrakingDec}
	\end{center}
\vspace{-1cm}
\end{table}
%
Current research is more focused on classifying the deceleration models and refining the deceleration characteristics for different functions and different scenarios, e.g. ACC-equipped cars. The deceleration and acceleration gradient rate are strictly limited to meet the requirements of comfort function as defined in \cite{ISO22179}. Unlike the ACC functional features, the driver braking reacts differently to complex traffic conditions and has more variable deceleration characteristics. With the electrification of vehicles increasing and the traffic environment becoming more complex, different deceleration models need to be classified so that they can be more relevant to the actual traffic conditions in the simulation. In general, the average deceleration will be based on experimental data, taking into account different driving styles and scenarios. A general conclusion can be drawn from the experimental statistics under normal driving conditions, the deceleration rate will not exceed -3.5 m/s$^{2}$, refer to \cite{Xu}. To refine the scenario, traffic environment and acceleration data for different drivers, the deceleration and braking process is represented using a parabolic model from \cite{Deligianni} combined with a maximum deceleration table, where the deceleration parameters can be based on different factors (e.g. gender, traffic flow, traffic conditions, age, etc.). This provides a reference to calibrate deceleration rate, braking distance, and duration time. Deceleration behavior model is divided into two categories. Tab. \ref{tab:DriverBrakingDec} shows the driver braking model which is used to simulate the braking characteristics of a real driver, and Tab. \ref{tab:ACCrBrakingDec} shows the ACC braking model which is used to simulate the deceleration profile with ACC driving function activated.

\subsubsection{Driver Braking Behavior}

Past studies about driver behavior proposed that vehicle needs more time and distance to decelerate at high speed. In this sense, the vehicle deceleration rate at high speed is much higher than at low speed. In order to cooperate with the simulation of vehicle dynamic characteristics at high speed, the deceleration rate is determined according to the speed range and target speed as an input condition which is shown in Tab. \ref{tab:bigTable}. 
\begin{table}[H]
	\resizebox{8.5cm}{!} {
	
		 	\renewcommand{\arraystretch}{1}
			\begin{tabular}{|p{0.05\textwidth}<{\centering}|p{0.05\textwidth}<{\centering} p{0.05\textwidth}<{\centering}|p{0.05\textwidth}<{\centering} p{0.05\textwidth}<{\centering}|p{0.05\textwidth}<{\centering} p{0.05\textwidth}<{\centering}|p{0.05\textwidth}<{\centering} p{0.05\textwidth}<{\centering}|p{0.05\textwidth}<{\centering} p{0.05\textwidth}<{\centering}|}
				\hline				
				\hline
				& \multicolumn{10}{c|}{Approach Speed}\\
				\cline{2-11}
				 & \multicolumn{2}{c|}{40-50} & \multicolumn{2}{c|}{50-60} & \multicolumn{2}{c|}{60-70} & \multicolumn{2}{c|}{70-80} & \multicolumn{2}{c|}{80-90}\\
				\cline{2-3} \cline{4-5} \cline{6-7} \cline{8-9} \cline{10-11}
				Speed  & speed & Decel & speed & Decel & speed & Decel & speed & Decel & speed & Decel \\
				interval & (km/h) & (m/s$^{2}$) 	& (km/h) & (m/s$^{2}$) 	& (km/h) & (m/s$^{2}$) 	& (km/h) & (m/s$^{2}$) 	& (km/h) & (m/s$^{2}$)\\
				(km/h)&&&&&&&&&&\\
				\hline
				0-10  &  2.64 & 0.91 &  2.81 & 0.84 &  2.43 & 0.88 &  2.62 & 0.89 &   2.6 & 0.87\\
				10-20 & 14.95 & 1.92 & 14.91 & 1.87 & 14.82 & 1.91 & 14.95 &    2 & 14.77 & 1.9\\
				20-30 &  25.1 & 1.82 & 25.03 & 1.92 & 25.07 & 2.12 & 24.84 & 1.71 & 24.87 & 2.07\\
				30-40 & 35.46 & 1.26 & 35.32 & 1.67 &  35.2 & 2.06 & 35.29 & 1.83 & 34.57 & 2.12\\
				40-50 & 44.04 & 0.67 & 45.46 &  1.1 & 45.23 & 1.75 & 44.83 & 1.76 & 45.12 & 2.02\\
				50-60 &   -    &   -   & 54.01 & 0.58 & 55.41 & 1.07 & 55.16 & 1.37 &  55.3 & 1.83\\
				60-70 &   -    &   -   &   -    &  -    & 63.69 & 0.58 & 65.49 & 0.78 &  65.2 & 1.34\\
				70-80 &   -    &   -    &   -   &  -    &   -    &   -   & 72.88 & 0.45 & 75.86 & 0.91\\
				80-90 &   -    &   -   &   -    &  -    &   -    &   -   &   -    &  -    &  82.9 & 0.48\\
				\hline
				\hline				
			\end{tabular}
		}
		\caption{Average deceleration rates by approach speed and speed during deceleration, table source \cite{Wang}. \hspace{\textwidth}}
		\label{tab:bigTable}
		\vspace{-0.5cm}
\end{table}
%
\begin{figure}[H]
	\centering
	\begin{tikzpicture}
	\begin{groupplot}
	[
	group style={
		group size=1 by 2,
		vertical sep=0.5cm,
		x descriptions at= edge bottom,
	},
	xmin = 1.5,
	xmax = 14,
	ylabsh=-2.5em,
	]
	\nextgroupplot
	[
	ymin= -2,   ymax=0.5,
	extra tick style={grid=major},
	width=0.9\linewidth,
	height=0.15\textheight,
	grid,
	ylabel style = {align = center},
	ylabel = Deceleration in m/s$^2$,
	title style={at={([yshift=-8ex]0.5,0)},anchor=north}, 
	subtitle/.style={title=\gpsubtitle{#1}},
	]
	\addplot[color=blue, mark = ., line width=1pt] table[x, y] {data/Driver_accel_t.txt};
	\nextgroupplot
	[
	ymin= 20,   ymax=80,
	extra tick style={grid=major},
	width=0.9\linewidth,
	height=0.15\textheight,
	grid,
	xlabel = {Time in s},
	ylabel style = {align = center},
	ylabel = Speed in m/s,
	yticklabel style={
		/pgf/number format/fixed,
		/pgf/number format/precision=5
	},
	title style={at={([yshift=-8ex]0.5,0)},anchor=north}, 
	subtitle/.style={title=\gpsubtitle{#1}},
	]
	\addplot[color=blue, mark = ., line width=1pt] table[x, y] {data/Driver_speed_t.txt};
	\end{groupplot}
	\end{tikzpicture}
	\caption{Driver braking relationship between deceleration rate and speed.}
	\label{fig:DriverBrakingSimulation}
\end{figure}
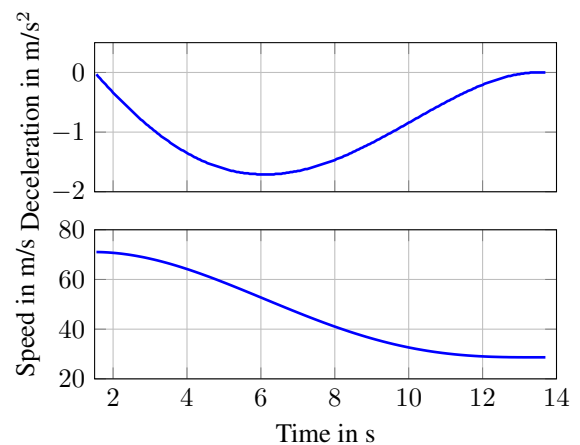
%
The driver braking model using external driver DLL interface is integrated into Vissim for testing. Fig. \ref{fig:DriverBrakingSimulation} shows the acceleration rate and vehicle velocity based on the polynomial model for a vehicle decelerating from the initial velocity 71.03 km/h to final velocity $v_{final}$ = 28.67 km/h. 
The deceleration time was set to 12 s and the maximum deceleration at -1.71 m/s$^{2}$.The results of Vissim simulation allow the observation of a slight oscillation around a constant velocity profile, which also corresponds to the real performance of the velocity behavior. The profile of driver braking model will drop rapidly to maximum deceleration in the beginning phase and then gradually return to 0 with the velocity decreasing.

\subsubsection{ACC Braking Behavior}

\begin{figure}[H]
	\centering
	\begin{tikzpicture}
	\begin{axis}
	[
		xmin=1.5,   xmax=30,
		extra tick style={grid=major},
		width=0.9\linewidth,
		height=0.2\textheight,
		grid,
		xlabel = {Time in s},
		ylabel = {Deceleration in m/s$^{2}$},
		title style={at={([yshift=-8ex]0.5,0)},anchor=north}, 
		subtitle/.style={title=\gpsubtitle{#1}},
		legend columns=1,
		legend pos = south east,
	]
	\addplot[color=blue, mark = ., line width=1pt] table[x, y] {data/ISO22179_maximum_jerk_profile.txt};
	\end{axis}
	\end{tikzpicture}
	\caption{Maximum negative jerk.}
	\label{fig:maxJerk}
\end{figure}
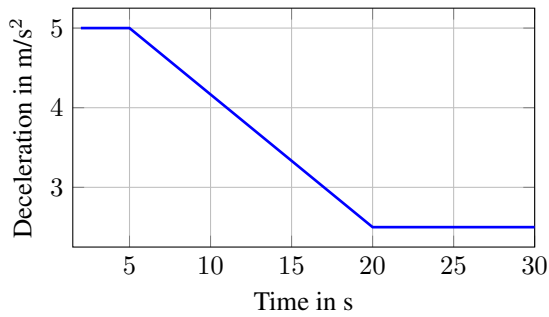

\begin{figure}
	\centering
	\begin{tikzpicture}
	\begin{groupplot}
	[
	group style={
		group size=1 by 2,
		vertical sep=0.5cm,
		x descriptions at= edge bottom,
	},
	xmin = 1.5,
	xmax = 5.5,
	ylabsh=-2.5em,
	]
	\nextgroupplot
	[
	ymin= -4,   ymax=0.5,
	extra tick style={grid=major},
	width=0.9\linewidth,
	height=0.2\textheight,
	grid,
	ylabel style = {align = center},
	ylabel = Deceleration in m/s$^2$,
	title style={at={([yshift=-8ex]0.5,0)},anchor=north}, 
	subtitle/.style={title=\gpsubtitle{#1}},
	]
	\addplot[color=blue, mark = ., line width=1pt] table[x, y] {data/ACC_accel_t.txt};
	\nextgroupplot
	[
	ymin= 40,   ymax=75,
	extra tick style={grid=major},
	width=0.9\linewidth,
	height=0.2\textheight,
	grid,
	xlabel = {Zeit $t$ in s},
	ylabel style = {align = center},
	ylabel = Speed in m/s,
	yticklabel style={
		/pgf/number format/fixed,
		/pgf/number format/precision=5
	},
	title style={at={([yshift=-8ex]0.5,0)},anchor=north}, 
	subtitle/.style={title=\gpsubtitle{#1}},
	]
		\addplot[color=blue, mark = ., line width=1pt] table[x, y] {data/ACC_speed_t.txt};
	\end{groupplot}
	\end{tikzpicture}
	\caption{ACC braking relationship between deceleration rate and speed.}
	\label{fig:ACCBrakingSimulation}
\end{figure}
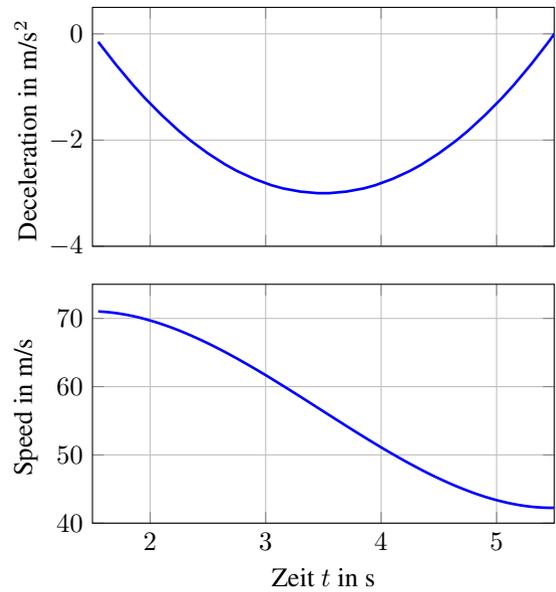

Since ACC is a comfortable longitudinal control function, the limitations of execution range in \cite{ISO22179} are used for the development and implementation of an ACC braking model. The deceleration gradient rate is limited in the standard, as shown in Fig. \ref{fig:maxJerk}.
%
 The average rate of deceleration gradient cannot exceed 2.5 m/s$^{3}$ when the speed is greater than 20 m/s and 5 m/s$^{3}$ when it is less than 5 m/s. The deceleration gradient is determined based on the initial vehicle speed. The entire deceleration process is expressed in the form of a parabolic as shown in Tab. \ref{tab:tableBrakingDec}.
\begin{table}[H]
	\begin{center}
			\renewcommand{\arraystretch}{1.5}
			\begin{tabular}{|c|l|l|}
				\hline
				\hline
				\multicolumn{3}{|c|}{\textbf{Parabolic Equation}}\\
				\hline

				\multicolumn{3}{|c|}{$y(t) = At^{2} + Bt +C$}\\
				
				\multicolumn{3}{|c|}{$A = (a_{0} - a_{1}) / \delta^{2}$}\\
				
				\multicolumn{3}{|c|}{$B = -2A\delta$}\\
				
				\multicolumn{3}{|c|}{$C = a_{0}$}\\
				\hline

				\multicolumn{3}{|c|}{\textbf{Parameters Description}}\\
				\hline
				\textbf{Variable} & \textbf{Comments} & \textbf{Unit}\\
				\hline
				t & Time variable & s\\
				\hline
				a$_{0}$ & Initial deceleration rate& m/s$^{2}$\\
				\hline
				a$_{1}$ & Final deceleration rate& m/s$^{2}$\\
				\hline
				y(t) & Deceleration rate& m/s$^{2}$\\
				\hline
				$\delta$ & Allowed maximum deceleration gradient rate & m/$s^{3}$\\
				\hline
				\hline				
			\end{tabular}	
		\caption{ACC braking deceleration model. \hspace{\textwidth}}
		\label{tab:tableBrakingDec}
	\end{center}
\vspace{-1cm}
\end{table}
%
The implemented ACC braking model in Vissim as depicted in Fig. \ref{fig:ACCBrakingSimulation} shows that the profile is determined based on the parabolic model. Jerk value is set to -1.5 m/s$^{3}$ and maximum deceleration to -3 m/s$^{2}$  at the time of simulation. The initial velocity is reduced from 70.97 km/h to 42.25 km/h considering the maximum deceleration gradient rate. In comparison with driver braking model, the deceleration gradient becomes a key factor to limit ACC deceleration profile. Therefore, the whole braking process is derived by a parabolic model.
%

\subsection{Stress Testing of Lateral Scenarios}\label{sec:latBehavior}
\begin{figure}[H]
	\centering
	\scalebox{0.8}{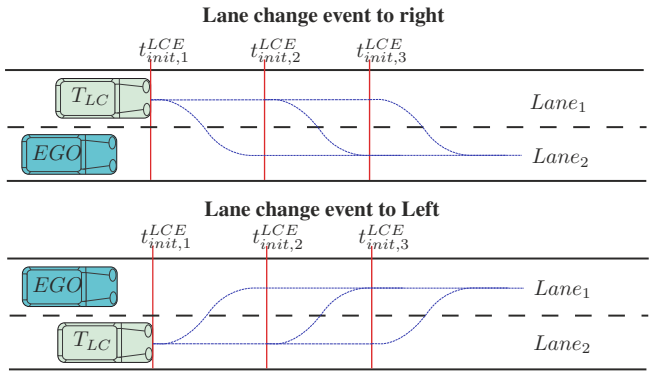}
	\caption{Relevant lane change events for 2 lane scenarios.}
	\label{fig:2lane_LC}
\end{figure}

\begin{figure}
	\centering
	\scalebox{0.8}{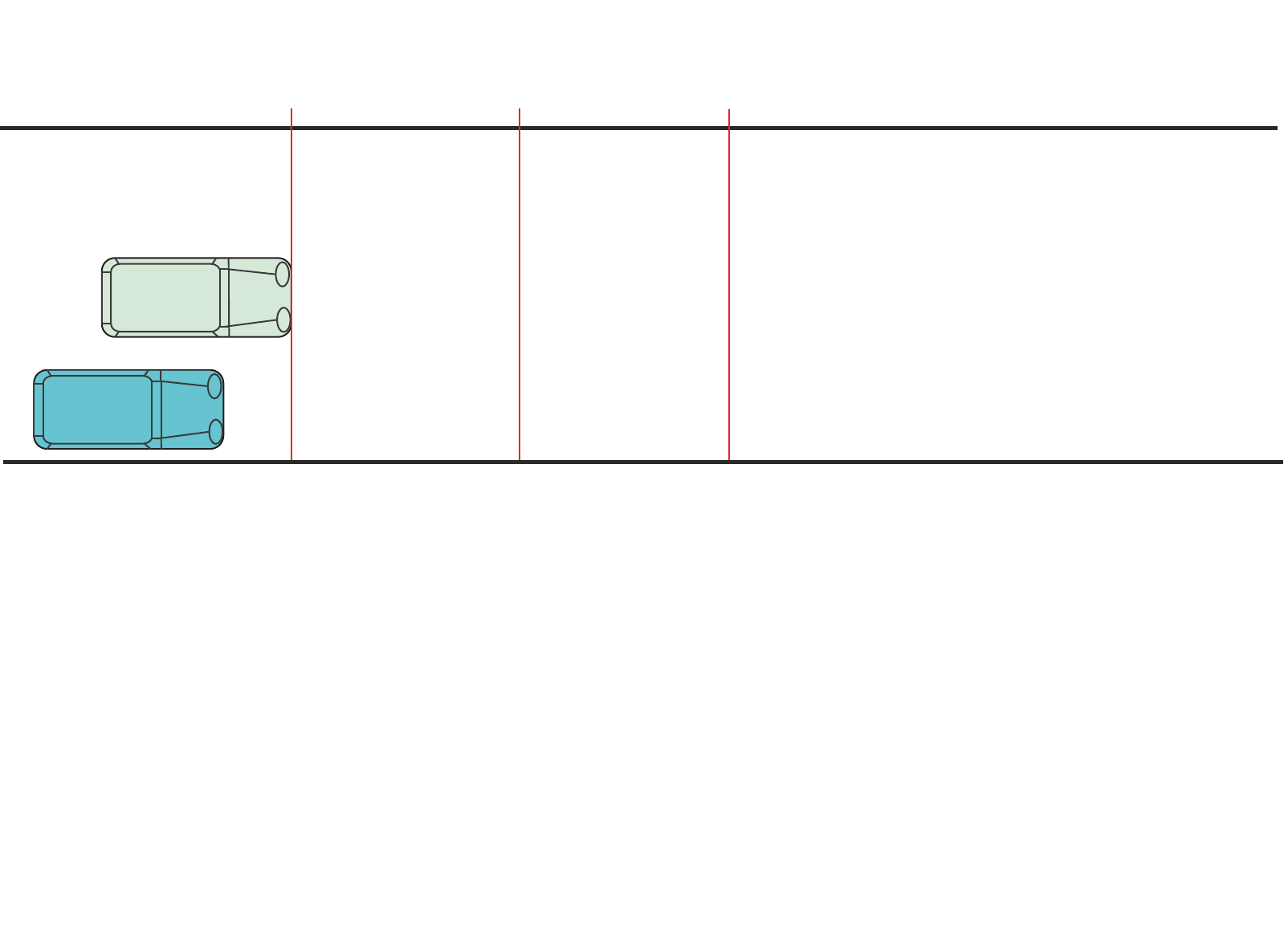}
	\caption{Relevant lane change events for 3 lane scenarios.}
	\label{fig:3lane_LC}
\end{figure}

\begin{figure}[H]
	\centering
	\scalebox{0.5}{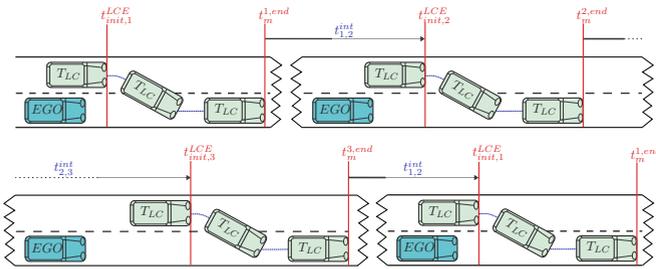}
	\caption{Description of the lane change event occurrence within each simulation testrun.}
	\label{fig:lceInterval}
\end{figure}

The second type of scenario considered for the STM are the lateral movements of traffic participants. As seen in Tab. \ref{tab:accidents}, accidents caused by unappropriated lateral behavior of traffic participants are second dominant in the data base of accidents. These accidents occur while a traffic participant initiates an aggressive cut-in maneuver from the left or right sight of a certain vehicle on road. In our case, the ego vehicle drives through the traffic and in case a traffic participant is overtaking or driving past the ego vehicle on the left or right lane. The cut-in maneuvers which are handled and manipulated by the STM are depicted in Fig. \ref{fig:2lane_LC} and \ref{fig:3lane_LC}. The times $t_{init,i}^{LCE}$ with $i\in\{1,2,3\}$ from Fig. \ref{fig:2lane_LC} and \ref{fig:3lane_LC} represent the initialization times of an aggressive lane change events (LCE). An LCE starts with initialization time $t_{init,i}^{LCE}$ and ends with the maneuver time $t_{m}$. During the simulation, each lane change event is executed one after the other with interval time $t_{i,j}^{int}$ between two lane change events. The minimum interval $t_{1,2}^{int}$  is set to 5 minutes. After these 5 minutes the next LCE is executed only in case a scenario depicted in Fig. \ref{fig:2lane_LC} and Fig. \ref{fig:3lane_LC} occur. After the last initialization time $t_{init,3}^{LCE}$ the events occur again from the beginning $t_{init,1}^{LCE}$. The description of this is shown in Fig. \ref{fig:lceInterval}. The initialization times $t_{init,i}^{LCE}$ and the event times $t_{1,2}^{int}$ are variables and can be adjusted within the Vissim interface. The lane change behavior and maneuver itself is described in the next subsection.
%
%
%

\subsection{Lane Change Behavior}
The implemented lane change behavior in Vissim is based on the lane change algorithm described in \cite{You} and \cite{Samiee}. Based on this research, the trajectory equation for the lane change can be derived in the displacement of the vehicle from the center of the lane in terms of time. The polynomial and linear equations from \ref{tab:LaneChangeEq} are used to determine the lateral $y_{lat}(t)$ and longitudinal $y_{long}(t)$ trajectory, respectively. This ensures a smooth trajectory and requires only a small number of points to generate the trajectory. In order to implement this algorithm in Vissim, a time-based lateral displacement equation, a time-based longitudinal displacement equation and acceleration profile are calculated and described in Tab. \ref{tab:LaneChangeEq}. To match more realistic lateral scenarios, the acceleration profile is calibrated with experimental data used in \cite{Samiee}. The entire process of lane change has an acceleration at the beginning phase and a gradual decrease in speed after the maneuver is completed. As a result, a sinusoidal function is involved to present lane change acceleration profile and shown in Tab. \ref{tab:LaneChangeEq}. 
\begin{table}[H]
	\begin{center}
		{
			\renewcommand{\arraystretch}{2}
			\centering
			\begin{tabular}{|P{1.8cm}|P{5.8cm}|}
				\hline
				\hline
				\multicolumn{2}{|c|}{\textbf{Lane change trajectory equation}}\\
				\hline
				Lateral trajectory & y$_{lat}$(t) = $\left(\dfrac{-6h}{t_{m}^{5}}\right)$t$^{5}$+$\left(\dfrac{15h}{t_{m}^{4}}\right)$t$^{4}$+$\left(\dfrac{-10h}{t_{m}^{3}}\right)$t$^{3}$\\
				\hline
				Longitudinal trajectory & y$_{long}$(t) = $v_{m}$t\\
				\hline
				Acceleration profile & a(t) = $a_{max}\sin\left(\dfrac{2\pi}{t_{m}}t\right)$\\
				\hline
				\multicolumn{2}{|c|}{\textbf{Parameters description}}\\
				\hline
				y$_{lat}$ & Lateral displacement\\
				\hline
				y$_{long}$ & Longitudinal displacement\\
				\hline
				$t_{m}$ & Maneuver time\\
				\hline
				$h$ & Maximum lateral displacement of the vehicle at the end of the maneuver\\
				\hline
				$v_{m}$ & Mean velocity during lane change maneuver\\
				\hline
				$a_{max}$ & Maximum acceleration value during lane change maneuver\\
				\hline
				\hline
			\end{tabular}
		}
		\caption{Trajectory parametrization of the lane change maneuver. \hspace{\textwidth}}
		\label{tab:LaneChangeEq}
	\end{center}
 	\vspace{-0.4cm}
\end{table}
The result calculated from lateral and longitudinal trajectory equations will be used as a reference to verify that Vissim output can follow the input commands. As an example, for a possible lane change scenario, a lane change maneuver is built with the longitudinal vehicle velocity $v_{m}$= 85 km/h. The total lane change maneuver time is set to $t_{m}$= 6 s, which corresponds to an average maneuver time according to \cite{branko}. The lateral displacement $y_{lat}(t)$ in the Vissim road model from \cite{ftgFramework} is h=3.5 m. This represents the distance from one center line to another. Regarding the longitudinal behavior, the maximum acceleration value is set to $a_{max}$=1.2m/s$^{2}$. The trajectory equations for this example can be obtained in (\ref{eq:eq9}), (\ref{eq:eq10}) and (\ref{eq:eq11}).
\begin{align}
	y_{lat}(t) &= -0.0067t^{5}+0.0840t^{4}-0.2800t^{3} \label{eq:eq9}\\
	y_{long}(t) &= 85t \label{eq:eq10}\\
	a(t) &= 1.2\sin\left(\frac{2\pi}{6}t \right) \label{eq:eq11} 
\end{align}
The comparison between the calculated values $y_{lat}$ to the values $y_{lat}^{V}$ set by Vissim using the DLL is shown in Fig. \ref{fig:LaneChangeComparison}. A minor deviation from the desired values can be observed. The deviations are caused by Vissim internal cycle times (min. program cycle 0.5s) and settling times which cannot be manipulated. For the STM only the cut-in in front of the ego vehicle is important for the testing process. After a target vehicle finishes the lane change, the driver behavior is not controlled by the DLL. The traffic vehicles continue to drive on the road using internal Vissim driver models.
\pgfplotsset{every axis legend/.append style={at={(0.5,-0.65)},anchor=south}}
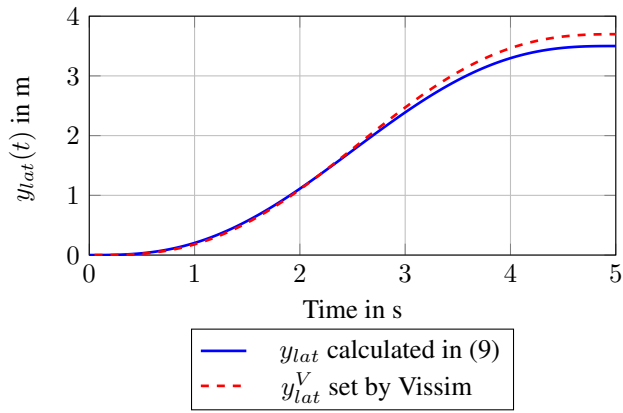
\begin{figure} [H]
	\centering
	\begin{tikzpicture}
	\begin{groupplot}
	[
	group style={
		group size=1 by 1,
		vertical sep=0.5cm,
		x descriptions at= edge bottom,
	},
	xmin = 0,
	xmax = 5,
	ylabsh=-2.5em,
	]
	\nextgroupplot
	[
		ymin= 0,   ymax=4,
		extra tick style={grid=major},
		width=1\linewidth,
		height=0.2\textheight,
		grid,
		xlabel = {Time in s},
		ylabel style = {align = center},
		ylabel = $y_{lat}(t)$ in m,
		yticklabel style={
			/pgf/number format/fixed,
			/pgf/number format/precision=5
	},
	title style={at={([yshift=-8ex]0.5,0)},anchor=north}, 
	subtitle/.style={title=\gpsubtitle{#1}},
	]
	\addplot[color=blue, mark = ., line width=1pt] table[x, y] {data/lat_pos_DLL.txt};
	\addlegendentry{\,\,\,\,\,\,$y_{lat}$ calculated in (\ref{eq:eq9})}
	\addplot[color=red,dashed, mark = ., line width=1pt] table[x, y] {data/lat_pos_Vissim.txt};
	\addlegendentry{$y_{lat}^{V}$ set by Vissim}
	\end{groupplot}
	\end{tikzpicture}
	\caption{Comparison between the desired signal generated by the DLL and the signal actually produced by Vissim.}
	\label{fig:LaneChangeComparison}
\end{figure}

\section{RESULTS}\label{sec:results}
To demonstrate the effectiveness of the STM, the co-simulation framework developed in \cite{ftgFramework} is used. The ego vehicle is an internal IPG Driver which is equipped with an ACC and automatic lane change algorithm. For the increase of the number of scenarios relevant for testing of ADS, the Vissim traffic is featured with the STM using an external driver model DLL interface. For this purpose, a Driver Model framework was developed and presented in \cite{deminDLL} to manipulate traffic vehicles in Vissim in order to provoke critical scenarios. Therefore, two simulation runs with 5000 kilometers are carried out. The first simulation run is carried out without the manipulation procedure and the second simulation run contains the manipulation according the presented STM. Critical scenarios which should be detected and evaluated, collisions and critical scenarios as defined in \cite{junitz} are taken to prove the efficiency of the STM. Based on the data from \cite{junitz} time-to-brake (TTB) and the requested acceleration are used to assess the detected scenarios. In \cite{junitz} three criticality levels are defined, non-critical, eventually critical and very critical. One possible and exemplary STM parameter configuration chosen by the engineer judgment for this research is defined in Tab. \ref{tab:stmConfig}. Tab. \ref{tab:stmResult} shows that the number of detected collisions increased from 59 to 625 on 5000 km. According to the metrics defined in \cite{junitz} very critical and eventually critical scenarios increase significantly compared to the co-simulation without the STM. Other parameter configurations are allowed and will be used for parameter studies in future research.
%
\begin{table}
	\centering
	\resizebox{0.475\textwidth}{!}{
		\begin{tabular}{|l|c|l|c|}
			\hline \hline
			\multicolumn{2}{|c|}{\specialcell{\textbf{Configuration} \\ \textbf{for longitudinal STM}}} & \multicolumn{2}{|c|}{\specialcell{\textbf{Parameter Configuration}\\ \textbf{for lateral STM}}} \\ \hline
			Parameter & Value        & Parameter & Value   \\ \hline
			$t_{1}^{s}$                     & 2 s          & $t_{m}$                        & 6 s      \\ \hline
			$t_{2}^{s}$                     & 4 s          & $a_{max}$                     & 1.2 m/s$^2$ \\ \hline
			$t_{3}^{s}$                     & 6 s          &    			     \\ \cline{1-2}
			$v_{final}$                     & 20 km/h      &                     \\ \cline{1-2}
			$t_{d}$                         & 12 s         &                     \\ \cline{1-2}
			$t_{d,max}$                     & -1.7 m/s$^2$ &                     \\ \hline \hline
		\end{tabular}
	}
	\caption{Configuration of the longitudinal driver braking behaviour and lateral behavior of the STM.}
	\label{tab:stmConfig}
\end{table}

%
%
\begin{table}
	\centering
	\resizebox{0.475\textwidth}{!}{%
		\begin{tabular}{|c|c|c|}
			\hline \hline
			\textbf{Scenario} & \specialcell{\textbf{Number of detections} \\ \textbf{without STM}} & \specialcell{\textbf{Number of detections} \\\textbf{with STM}} \\ \hline
			Collisions & None & 625 \\ \hline
			\multicolumn{3}{|c|}{Scenarios based on Evaluation Metric from \cite{junitz}} \\ \hline
			Eventually Critical & 937 & 3257 \\ \hline
			Very Critical & 298 & 2157 \\ \hline \hline
		\end{tabular}%
	}
	\caption{Comparison of simulation results with and without the STM.}
	\label{tab:stmResult}
\end{table}
%

\section{CONCLUSIONS}
In this paper we presented a stress testing method to increase and generate safety critical scenarios in a complex co-simulation between IPG CarMaker and PTV Vissim. Based on statistical accident data for motorways in Austria,  two dominant accident types of scenarios are extracted and used for the presented method. Using these accident types, defined maneuvers are provoked by traffic participants in the surrounding area of the vehicle under test. The developed stress testing method showed a significant increase of detected scenarios in the co-simulation environment and will be used for further research and development. In addition to manipulating traffic vehicles on the motorway, a huge potential using the presented method lies in the use for urban areas where the environment is far more complex than on motorways.

\begin{IEEEbiography}[{\includegraphics[width=1in,height=1.25in,clip,keepaspectratio]{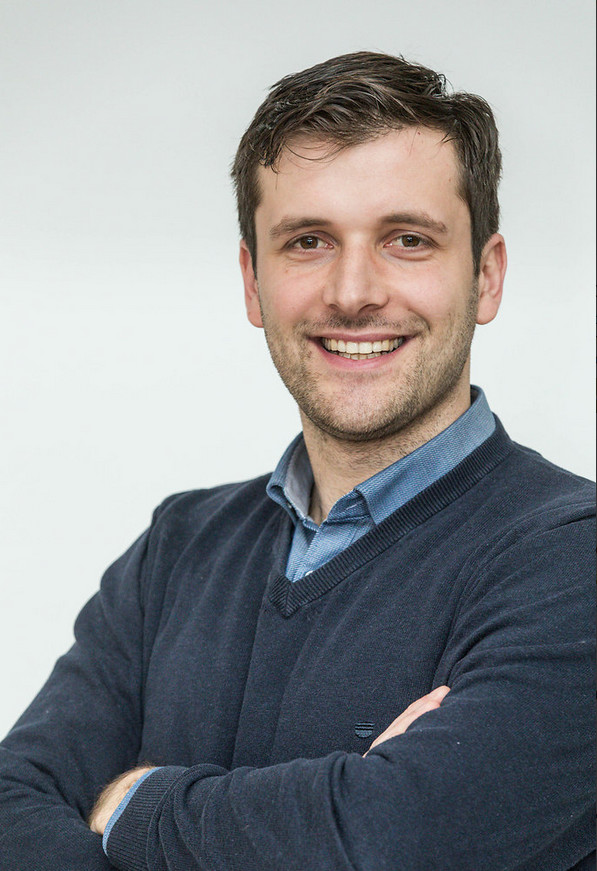}}]{Demin Nalic} received his bachelor degree from Vienna University of Technology in Electrical Engineering and Information Technologies and his Master’s degree in Control and Automation Engineering, 2016 from the same University. He was software engineer and project leader for smart manufacturing in Siemens Austria from 2016 to 2018. Since 2019 he is member of IEEE and within IEEE member of the Intelligent Transportation Systems Society. Currently he is working as research assistant at the institute of automotive engineering at the Graz University of Technology. His research interests include autonomous systems, testing methodologies and control solutions for automated driving systems.
\end{IEEEbiography}
\vspace{0cm}
\begin{IEEEbiography}[{\includegraphics[width=1in,height=1.25in,clip,keepaspectratio]{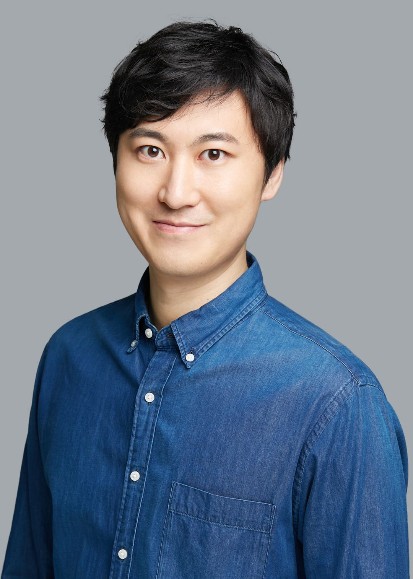}}]{Hexuan Li} received the B.S. and M.S degree from University of Clermont Auvergne, France, 2016, both in mechatronics engineering.	From 2016 to 2020, he was employed at PSA and Audi China successively, mainly responsible for the highly automated driving system integration and hardware-in-the-loop simulation. Since 2020, he has been working with the Institute of Automotive Engineering, University of Technology Graz, dealing with Driver Assistance Systems, sensor modelling and co-simulation framework based on a full vehicle test bench.	
\end{IEEEbiography}
\vspace{0cm}
\begin{IEEEbiography}[{\includegraphics[width=1in,height=1.25in,clip,keepaspectratio]{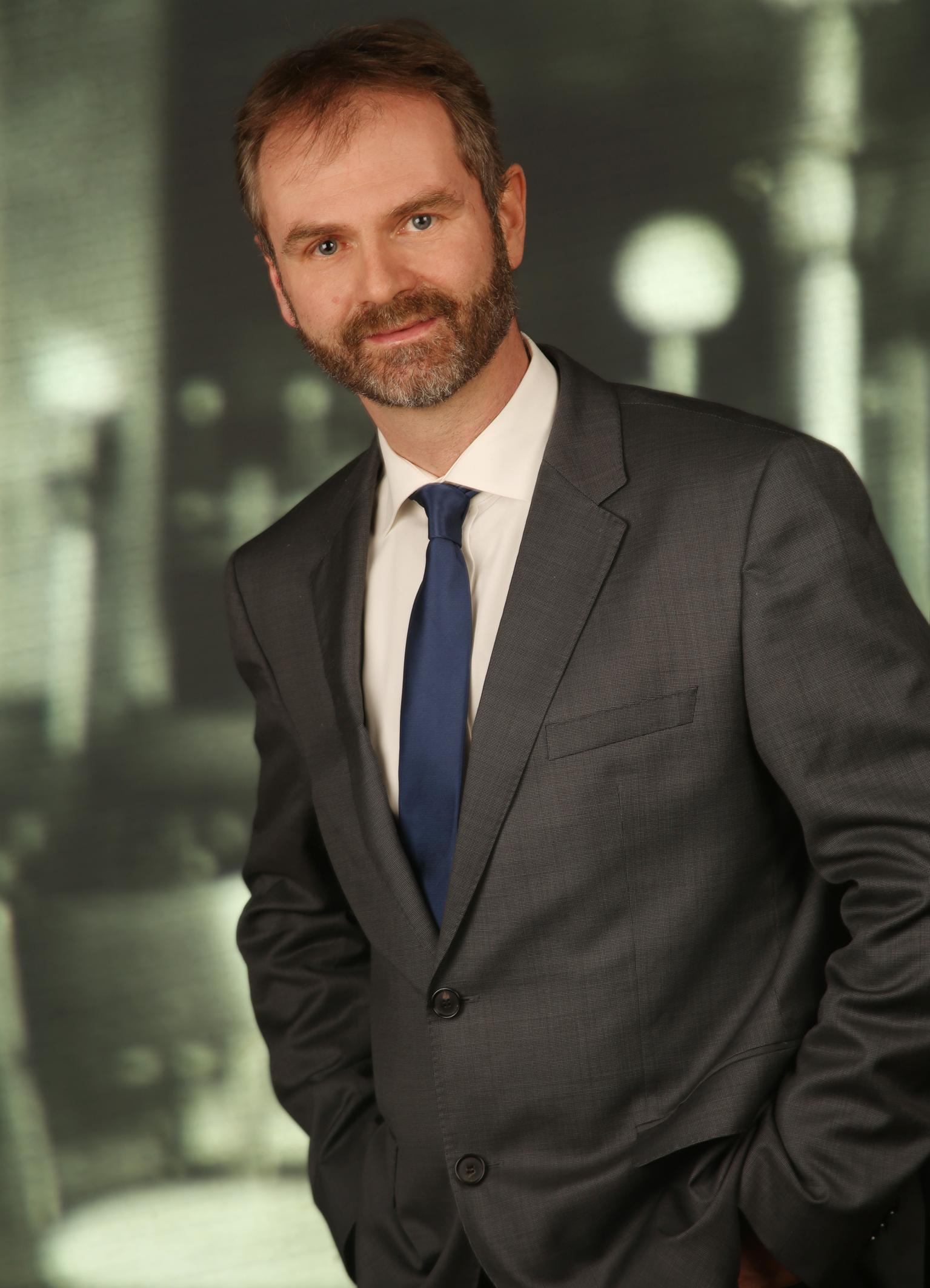}}]{Arno Eichberger} received the degree in mechanical engineering from the University of Technology Graz, in 1995, and the Ph.D. degree (Hons.) in technical sciences, in 1998. From 1998 to 2007, he was employed at MAGNA STEYR Fahrzeugtechnik AG\&Co and dealt with different aspects of active and passive safety. Since2007, he has been working with the Institute of Automotive Engineering, University of Technology Graz, dealing with Driver Assistance Systems,Vehicle Dynamics and Suspensions. Since 2012, he has been an Associate Professor holding a venia docendi of automotive engineering.
\end{IEEEbiography}
\vspace{0cm}
\begin{IEEEbiography}[{\includegraphics[width=1in,height=1.25in,clip,keepaspectratio]{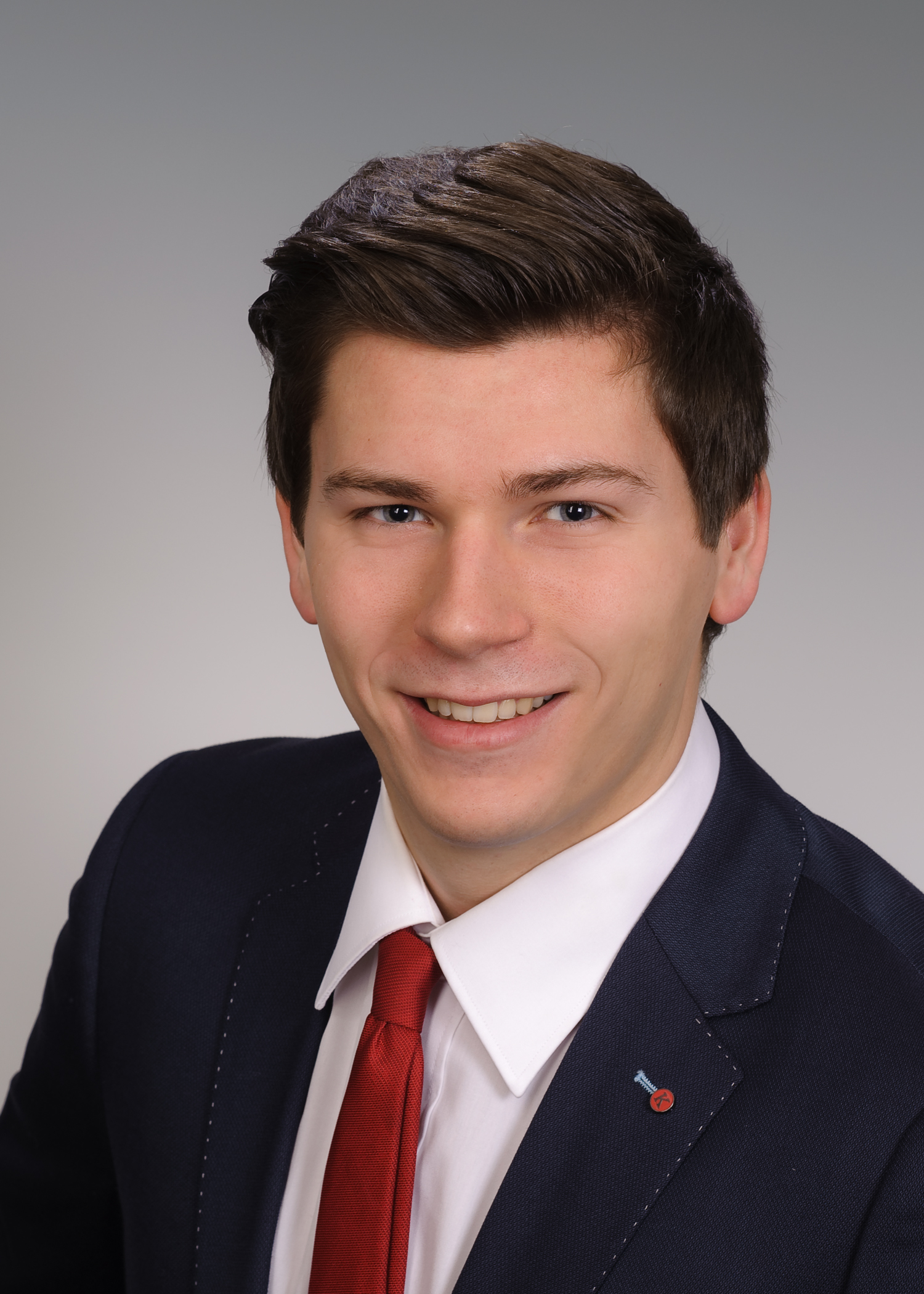}}]{Christoph Wellershaus} received the BSc degree in mechanical engineering and business economics from Technical University of Graz, Austria, in 2019. He is currently pursuing his MSc degree in mechanical engineering and business economics at Technical University of Graz, Austria. He started to work as a scientific student assistant in the research field of driver assistance and automated driving at the Institute of Automotive Engineering at Technical University of Graz in 2018. From 2016 to 2018 he participated in the Formula Student - student association “TU Graz Racing Team”, where he held the role of the team leader from 2017 to 2018. 
\end{IEEEbiography}
\vspace{-0cm}
\begin{IEEEbiography}[{\includegraphics[width=1in,height=1.25in,clip,keepaspectratio]{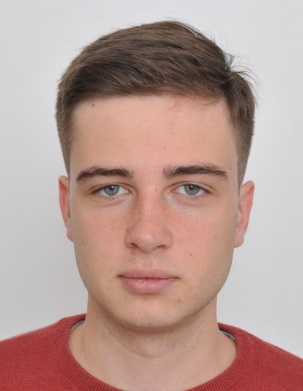}}]{Aleksa Pandurevic} received the BSc degree in computer science from the Graz University of Technology, Graz, Austria, in 2020. Currently, he is studying for a MSc computer science degree at the Graz University of Technology and Technical University of Munich.	In 2019, he joined the Institute of Automotive Engineering at Graz University of Technology as a Student assistant on scientific projects where he contributed to different projects in field of automated driving and driver assistance systems. Since 2020, he also works as a software development intern at Infineon Technologies Austria AG, Graz. His current topics of interest are machine learning, data science and information security.
\end{IEEEbiography}
\vspace{-0cm}
\begin{IEEEbiography}[{\includegraphics[width=1in,height=1.25in,clip,keepaspectratio]{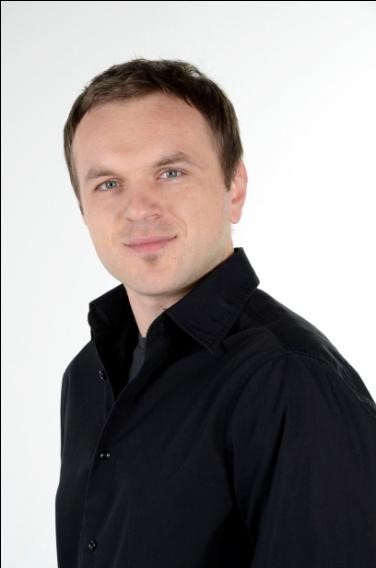}}]{Branko Rogic} recived his bachelor and master degree in mechanical engineering from the Technical University of Graz. His final degree project dealt with the simulation of hybrid commercial vehicle drivetrains and was carried out on the Institute of Automotive Engineering. After the graduation he worked as a development engineer at the automotive company ZF (in 2013 and 2014). 2015 he returned to the Institute of Automotive Engineering to work on an industrial research project, dealing with the integration of ADAS functions and in cooperation with Magna Steyr. Since 2018 he is an employee by Magna Steyr at the ADAS advanced development department.
\end{IEEEbiography}

\EOD


\begin{thebibliography}{99}

\bibitem{padock} Kalra, Nidhi \& Paddock, Susan. (2016). Driving to safety: How many miles of driving would it take to demonstrate autonomous vehicle reliability?. Transportation Research Part A: Policy and Practice. 94. 182-193. 10.1016/j.tra.2016.09.010. 

\bibitem{winner} W. Wachenfeld,H. Winner, Die Freigabe des autonomen Fahrens.In Lenz B, Winner H, Gerdes JC, Maurer M editors. Autonomes Fahren: technische, rechtliche und gesellschaftliche Aspekte, Vol. 116.s.l. Heidelberg, Germany: Springer. p. 439–464.

\bibitem{survey1} Felix Batsch, Stratis Kanarachos, Madeline Cheah, Roberto Ponticelli and Mike Blundell, "A taxonomy of validation strategies to ensure the safe operation of highly automated vehicles," Journal of Intelligent Transportation Systems, DOI: 10.1080/15472450.2020.1738231

\bibitem{survey2} D. Nalic, T. Mihalj, M. Bäumler, M. Lehmann, A. Eichberger and S. Bernsteiner, "Scenario Based Testing of Automated Driving Systems: A Literature Survey," FISITA Web Congress, 2020, pp. xxx-xxx. 
\bibitem{critical_1} T. Mugur, "Enhancing ADAS Test and Validation with Automated Search for Critical Situations", presented at Driving Simulation Conference \& Exhibition 2015, Berlin, September 16-18, 2015.

\bibitem{ftgFramework} D. Nalic, A. Eichberger, G. Hanzl, M. Fellendorf and B. Rogic, "Development of a Co-Simulation Framework for Systematic Generation of Scenarios for Testing and Validation of Automated Driving Systems*," 2019 IEEE Intelligent Transportation Systems Conference (ITSC), Auckland, New Zealand, 2019, pp. 1895-1901, doi: 10.1109/ITSC.2019.8916839.

\bibitem{TrafficFlowTesting1} S. Hallerbach, Y. Xia, U. Eberle, and F. Koester, "Simulation-based identification  of  critical  scenarios  for  cooperative  and  automated vehicles", SAE  International  Journal  of  Connected  and  Automated Vehicles, vol. 1, no. 2018-01-1066, pp. 93–106, 2018.

\bibitem{TrafficFlowTesting2} T. Helmer, L. Wang, K. Kompass and R. Kates, "Safety Performance Assessment of Assisted and Automated Driving by Virtual Experiments: Stochastic Microscopic Traffic Simulation as Knowledge Synthesis," 2015 IEEE 18th International Conference on Intelligent Transportation Systems, Las Palmas, 2015, pp. 2019-2023, doi: 10.1109/ITSC.2015.327.

\bibitem{TrafficFlowTesting3}  D. Gruyer, S. Choi, C. Boussard and B. d'Andréa-Novel, "From virtual to reality, how to prototype, test and evaluate new ADAS: Application to automatic car parking," 2014 IEEE Intelligent Vehicles Symposium Proceedings, Dearborn, MI, 2014, pp. 261-267, doi: 10.1109/IVS.2014.6856525.

\bibitem{statisticAustria} Statistic Austria, Accessed at 01.12.2020. [online] Available: \url{https://www.statistik.at/web_en/statistics/index.html}

\bibitem{Vissim} M. Fellendorf and P. Vortisch,   "Microscopic   traffic   flow   simulator VISSIM," in Fundamentals of Traffic Simulation,  vol. 145, ser. International Series in Operations Research and Management Science,  J. Barcel,Ed.New York, NY, USA: Springer-Verlag, 2010, pp. 63–93.

\bibitem{TrafficFlow} J. Barceló, Ed.,Fundamentals of Traffic Simulation, vol. 145. New York,NY, USA: Springer,  2010.

\bibitem{Wang} B. Wang, W. Chen, B. Zhang, Y. Zhao, "Regulation cooperative control for heterogeneous uncertain chaotic systems with time delay: A synchronization errors estimation framework," Automatica, Vol. 108, 2019, https://doi.org/10.1016/j.automatica.2019.06.038.

\bibitem{Deng} C. Deng and C. Wen, "Distributed Resilient Observer-Based Fault-Tolerant Control for Heterogeneous Multiagent Systems Under Actuator Faults and DoS Attacks," in IEEE Transactions on Control of Network Systems, vol. 7, no. 3, pp. 1308-1318, Sept. 2020, doi: 10.1109/TCNS.2020.2972601.

\bibitem{ifacFramework} D. Nalic, A. Pandurevic, A., A. Eichberger and B. Rogic, "Design and Implementation of a Co-Simulation Framework for Testing of Automated Driving Systems,", Preprints 2020, 2020110252.
\bibitem{trapp} R. Trapp, "Hinweise zur mikroskopischen Verkehrsflusssimulation-Grundlagen und Anwendung", Forschungsgesellschaft für strassen-und verkehrswesen (2006), Köln: FGSV Verlag GmbH.
\bibitem{alplab} S. Seebacher, B. Datler, J. Erhart, M. Harrer, P. Hrassnig, A. Präsent, C. Schwarzl and M. Ullrich, "Infrastructure data fusion for validation and future enhancements of autonomous vehicles perception on Austrian motorways," IEEE International Conference on Connected Vehicles and Expo (ICCVE), Graz, Austria, 2019, pp. 1-7, doi: 10.1109/ICCVE45908.2019.8965142.
\bibitem{deminDLL} D. Nalic, A. Pandurevic, A. Eichberger, B. Rogic, "Testing of Automated Driving Systems in a Dynamic Traffic Environment", submitted to SoftwareX, [arXiv:cs.SE/2011.05798].
\bibitem{statisticAustriaAccidentClassification} Statistic Austria Accident Types, [online] Available:  \url{https://www.statistik.at/web_en/statistics/EnergyEnvironmentInnovationMobility/transport/road/road_traffic_accidents/index.html}
\bibitem{parIdent1} M. Klischat and M. Althoff, "Generating Critical Test Scenarios for Automated Vehicles with Evolutionary Algorithms," 2019 IEEE Intelligent Vehicles Symposium (IV), Paris, France, 2019, pp. 2352-2358, doi: 10.1109/IVS.2019.8814230.
\bibitem{parIdent2} M. R. Zofka, F. Kuhnt, R. Kohlhaas, C. Rist, T. Schamm and J. M. Zöllner, "Data-driven simulation and parametrization of traffic scenarios for the development of advanced driver assistance systems," 2015 18th International Conference on Information Fusion (Fusion), Washington, DC, 2015, pp. 1422-1428.
\bibitem{concreteScenario} T. Menzel, G. Bagschik and M. Maurer, "Scenarios for Development, Test and Validation of Automated Vehicles," 2018 IEEE Intelligent Vehicles Symposium (IV), Changshu, 2018, pp. 1821-1827, doi: 10.1109/IVS.2018.8500406.
\bibitem{activeSafety1} European New Car Assessment Program (Euro-NCAP). Frontal Impact Testing Protocol, Version 4.3. Testing  protocol, Euro-NCAP, February 2009.
\bibitem{activeSafety2} S. K. Gehrig and F. J. Stein, "Collision Avoidance for Vehicle-Following Systems," in IEEE Transactions on Intelligent Transportation Systems, vol. 8, no. 2, pp. 233-244, June 2007, doi: 10.1109/TITS.2006.888594.
\bibitem{activeSafety3} H. Winner, S. Hakuli and G. Wolf, "Handbuch Fahrerassistenzsysteme: Grundlagen, Komponenten und Systeme fur aktive Sicherheit und Komfort,",  2nd ed. (in German). Wiesbaden, Germany: Vieweg+Teubner, 2011.
\bibitem{activeSafety4} National Highway and Traffic Safety Administration (NHTSA).  Reportto Congress on the National Highway Traffic Safety Administration ITSProgram. Program Progress During 1992-1996 and Strategic Plan for 1997-2002.  Technical report, NHTSA, 1997.
\bibitem{activeSafety5} D. Wallner,  A. Eichberger,  and  W. Hirschberg. "A  Novel  Control  Al-gorithm  for  Integration  of  Active  and  Passive  Vehicle  Safety  Systemsin  Frontal  Collisions," In Proceedings of the 2nd International  Multi-Conference on Engineering and Technological Innovation (IMETI), 10-13 July 2009.  Orlando, USA.
\bibitem{Akcelik1} Akçelik, Rahmi and D. C. Biggs. "Acceleration profile models for vehicles in road traffic," Transportation Science, vol. 21, no. 1, pp: 36-54., 1987.
\bibitem{Maurya} A. K. Maurya and P. S. Bokare, "Study of deceleration behaviour of different vehicle types," International Journal for Traffic Transportation Engineering, vol. 2, no. 3, pp. 253–270, 2012.
\bibitem{Bokare} Bokare, P. S. and A. K. Maurya. "Acceleration-deceleration behaviour of various vehicle types," Transportation Research Procedia, vol. 5, pp: 4733-474, 2017.
\bibitem{Akcelik2} Akçelik, Rahmi and Mark Besley. "Acceleration and deceleration models," in 23nd Proceedings Conference of Australian Institutes of Transport Research (CAITR 2001), Vol. 10, pp: 1-9, Melbourne Australia, 2001.

\bibitem{Kudarauskas} Kudarauskas, Nerijus. "Analysis of emergency braking of a vehicle," Transport, vol. 22, no. 3 pp: 154-159, 2007.

\bibitem{ISO22179} A. Nix and J. Kemp, "Full speed range adaptive cruise control system," US Patent 20090254260A1, October, 8th, 2009.

\bibitem{Xu} Xu, Jin, et al. "Acceleration and deceleration calibration of operating speed prediction models for two-lane mountain highways," Journal of Transportation Engineering, Part A, Systems, vol. 143, no. 7, July 2017, Art. no. 04017024.

\bibitem{Deligianni} D.S. Panagiota, M. Quddus, A. Anvuur and S. Reed, "Analyzing and modeling drivers' deceleration behavior from normal driving," Transportation research record, Vol. 2663, no. 1, pp: 134-141, 2017.

\bibitem{Wang2} J. Wang, K. K. Dixon, H. Li and J. Ogle, "Normal deceleration behavior of passenger vehicles at stop sign–controlled intersections evaluated with in-vehicle Global Positioning System data," Transportation research record, Vol. 1937, no. 1, pp:120-127, 2005.

\bibitem{P} P. Tientrakool, Y. Ho and N. F. Maxemchuk, "Highway Capacity Benefits from Using Vehicle-to-Vehicle Communication and Sensors for Collision Avoidance," 2011 IEEE Vehicular Technology Conference (VTC Fall), San Francisco, CA, 2011, pp. 1-5, doi: 10.1109/VETECF.2011.6093130.

\bibitem{Mehmood} A. Mehmood and S. M. Easa, "Modeling reaction time in car-following behaviour based on human factors," International Journal of Applied Science, Engineering and Technology, vol. 5, no. 14, pp: 93-101, 2009.

\bibitem{You} F. You, R. Zhang,G. Lie, H. Wang, H. Wen and J. Xu, "Trajectory planning and tracking control for autonomous lane change maneuver based on the cooperative vehicle infrastructure system,"  Expert Systems with Applications, vol. 42, no. 14, pp: 5932-5946, 2015.

\bibitem{Samiee} S. Samiee, S. Azadi, R. Kazemi and A. Eichberger "Towards a decision-making algorithm for automatic lane change manoeuvre considering traffic dynamics," PROMET-Traffic and Transportation, vol. 28, no. 2, pp: 91-103, 2016.

\bibitem{branko} B. Rogic, D. Nalic, A. Eichberger and S. Bernsteiner, "A Novel Approach to Integrate Human-in-the-Loop Testing in the Development Chain of Automated Driving: The Example of Automated Lane Change,", 21th IFAC World Congress, 2020.

\bibitem{junitz} P. Junietz, J. Schneider, H. Winner, "Metrik zur Bewertung der Kritikalität von Verkehrssituationen und -szenarien," (in German) in 11th Workshop Fahrerassistenzsysteme, 2017.

\end{thebibliography}
\end{document}